\documentclass{article}

\usepackage{styling/arxiv}

\usepackage[utf8]{inputenc} % allow utf-8 input
\usepackage[T1]{fontenc}    % use 8-bit T1 fonts
\usepackage{hyperref}       % hyperlinks
\usepackage{url}            % simple URL typesetting
\usepackage{booktabs}       % professional-quality tables
\usepackage{amsfonts}       % blackboard math symbols
\usepackage{nicefrac}       % compact symbols for 1/2, etc.
\usepackage{microtype}      % microtypography
\usepackage{lipsum}		% Can be removed after putting your text content
\usepackage{graphicx}
\usepackage{doi}
\usepackage{bm}
\usepackage{amsmath}
\usepackage{algorithm}
\usepackage{cite}
\usepackage{arydshln}
\usepackage{algpseudocode}
\usepackage{setspace}

\algnewcommand{\Inputs}[1]{%
  \State \textbf{inputs}
  \Statex \hspace*{\algorithmicindent}\parbox[t]{.8\linewidth}{\raggedright #1}
}

\newif\ifIEEE
\IEEEfalse

\usepackage[disable]{todonotes}   % add 'disable' to hide notes

\newcommand{\bart}[2][] {\todo[inline,backgroundcolor=green!20!white, #1]{(Bart) #2}}

\usepackage{pgfplots}
\pgfplotsset{compat=1.17}
\usetikzlibrary{intersections}
\usepackage{bm}
\usetikzlibrary{positioning}
\usepgfplotslibrary{fillbetween}
\usepgfplotslibrary{groupplots}
\usetikzlibrary{shapes.geometric,backgrounds}
\usepackage[outline]{contour}

\usepackage{tikz}
\usepackage{xifthen}

\pgfdeclarelayer{bg}    % declare background layer
\pgfsetlayers{bg,main}  % set the order of the layers (main is the standard layer)
\definecolor{beige}{RGB}{245, 245, 220}

\definecolor{darkgrey}{RGB}{75, 75, 75}
\definecolor{lightgrey}{RGB}{250, 250, 250}

\usetikzlibrary{calc, arrows, fit, positioning, patterns, decorations.pathreplacing, shapes}
\tikzstyle{dash} = [dashed, -latex,>=latex]
\tikzstyle{line} = [draw, -latex,>=latex]
\tikzstyle{smallbox} = [draw, minimum size=5.0mm]
\tikzstyle{box} = [draw, minimum size=7.0mm]
\tikzstyle{bigbox} = [draw, minimum size=10.0mm]
\tikzstyle{switch} = [trapezium, trapezium angle=120, draw, rotate=90,  inner ysep=5pt, outer sep=5pt,
minimum height=7mm, minimum width=7mm]
\tikzstyle{roundbox} = [draw, circle, inner sep=0pt, minimum size=3mm]
\tikzstyle{clamped} = [draw, fill=black, minimum size=0.15cm]
\tikzstyle{msgcircle} = [shape=circle, draw, inner sep=0pt, minimum size=4mm, fill=white, font=\scriptsize]
\tikzstyle{darkmsgcircle} = [shape=circle, draw, inner sep=0pt, minimum size=4mm, fill=darkgrey, text=white, font=\scriptsize]
\tikzstyle{msgdoublecircle} = [shape=circle, double, double distance=1.5pt, draw, inner sep=0pt, minimum size=5mm, fill=white]
\tikzstyle{darkmsgdoublecircle} = [shape=circle, double, double distance=1.5pt, draw, inner sep=0pt, minimum size=5mm, fill=darkgrey, text=white, font=\bfseries]
\newcommand{\msg}[6]{
      \ifthenelse{\isin{#1}{left} \AND \isin{#2}{down}}{
            \coordinate (anchor) at ($({#3})!{#5}!({#4})$);
            \node[xshift=-6.0mm] at (anchor) {#6};
            \node[xshift=-1.0mm] at (anchor) {$\downarrow$};
      }{}
      \ifthenelse{\isin{#1}{right} \AND \isin{#2}{down}}{
            \coordinate (anchor) at ($({#3})!{#5}!({#4})$);
            \node[xshift=6.0mm] at (anchor) {#6};
            \node[xshift=1.0mm] at (anchor) {$\downarrow$};
      }{}

      \ifthenelse{\isin{#1}{down} \AND \isin{#2}{right}}{
            \coordinate (anchor) at ($({#3})!{#5}!({#4})$);
            \node[ yshift=-4.0mm] at (anchor) {#6};
            \node[yshift=-1.0mm] at (anchor) {$\rightarrow$};
      }{}
      \ifthenelse{\isin{#1}{up} \AND \isin{#2}{right}}{
            \coordinate (anchor) at ($({#3})!{#5}!({#4})$);
            \node[ yshift=4.0mm] at (anchor) {#6};
            \node[yshift=1.0mm] at (anchor) {$\rightarrow$};
      }{}

      \ifthenelse{\isin{#1}{down} \AND \isin{#2}{left}}{
            \coordinate (anchor) at ($({#3})!{#5}!({#4})$);
            \node[ yshift=-4.0mm] at (anchor) {#6};
            \node[yshift=-1.0mm] at (anchor) {$\leftarrow$};
      }{}
      \ifthenelse{\isin{#1}{up} \AND \isin{#2}{left}}{
            \coordinate (anchor) at ($({#3})!{#5}!({#4})$);
            \node[ yshift=4.0mm] at (anchor) {#6};
            \node[yshift=1.0mm] at (anchor) {$\leftarrow$};
      }{}

      \ifthenelse{\isin{#1}{left} \AND \isin{#2}{up}}{
            \coordinate (anchor) at ($({#3})!{#5}!({#4})$);
            \node[ xshift=-6.0mm] at (anchor) {#6};
            \node[xshift=-1.0mm] at (anchor) {$\uparrow$};
      }{}
      \ifthenelse{\isin{#1}{right} \AND \isin{#2}{up}}{
            \coordinate (anchor) at ($({#3})!{#5}!({#4})$);
            \node[ xshift=6.0mm] at (anchor) {#6};
            \node[xshift=1.0mm] at (anchor) {$\uparrow$};
      }{}
}

\newcommand{\msgcircle}[6]{
      \ifthenelse{\isin{#1}{left} \AND \isin{#2}{down}}{
            \coordinate (anchor) at ($({#3})!{#5}!({#4})$);
            \node[msgcircle,xshift=-5.0mm] at (anchor) {#6};
            \node[xshift=-1.5mm] at (anchor) {$\downarrow$};
      }{}
      \ifthenelse{\isin{#1}{right} \AND \isin{#2}{down}}{
            \coordinate (anchor) at ($({#3})!{#5}!({#4})$);
            \node[msgcircle,xshift=5.0mm] at (anchor) {#6};
            \node[xshift=1.5mm] at (anchor) {$\downarrow$};
      }{}

      \ifthenelse{\isin{#1}{down} \AND \isin{#2}{right}}{
            \coordinate (anchor) at ($({#3})!{#5}!({#4})$);
            \node[msgcircle, yshift=-5.0mm] at (anchor) {#6};
            \node[yshift=-2.0mm] at (anchor) {$\rightarrow$};
      }{}
      \ifthenelse{\isin{#1}{up} \AND \isin{#2}{right}}{
            \coordinate (anchor) at ($({#3})!{#5}!({#4})$);
            \node[msgcircle, yshift=5.0mm] at (anchor) {#6};
            \node[yshift=2.0mm] at (anchor) {$\rightarrow$};
      }{}

      \ifthenelse{\isin{#1}{down} \AND \isin{#2}{left}}{
            \coordinate (anchor) at ($({#3})!{#5}!({#4})$);
            \node[msgcircle, yshift=-5.0mm] at (anchor) {#6};
            \node[yshift=-2.0mm] at (anchor) {$\leftarrow$};
      }{}
      \ifthenelse{\isin{#1}{up} \AND \isin{#2}{left}}{
            \coordinate (anchor) at ($({#3})!{#5}!({#4})$);
            \node[msgcircle, yshift=5.0mm] at (anchor) {#6};
            \node[yshift=2.0mm] at (anchor) {$\leftarrow$};
      }{}

      \ifthenelse{\isin{#1}{left} \AND \isin{#2}{up}}{
            \coordinate (anchor) at ($({#3})!{#5}!({#4})$);
            \node[msgcircle, xshift=-5.0mm] at (anchor) {#6};
            \node[xshift=-1.5mm] at (anchor) {$\uparrow$};
      }{}
      \ifthenelse{\isin{#1}{right} \AND \isin{#2}{up}}{
            \coordinate (anchor) at ($({#3})!{#5}!({#4})$);
            \node[msgcircle, xshift=5.0mm] at (anchor) {#6};
            \node[xshift=1.5mm] at (anchor) {$\uparrow$};
      }{}
}

\newcommand{\darkmsg}[6]{
      \ifthenelse{\isin{#1}{left} \AND \isin{#2}{down}}{
            \coordinate (anchor) at ($({#3})!{#5}!({#4})$);
            \node[darkmsgcircle, xshift=-5mm] at (anchor) {#6};
            \node[xshift=-1.5mm] at (anchor) {$\downarrow$};
      }{}
      \ifthenelse{\isin{#1}{right} \AND \isin{#2}{down}}{
            \coordinate (anchor) at ($({#3})!{#5}!({#4})$);
            \node[darkmsgcircle, xshift=5mm] at (anchor) {#6};
            \node[xshift=1.5mm] at (anchor) {$\downarrow$};
      }{}

      \ifthenelse{\isin{#1}{down} \AND \isin{#2}{right}}{
            \coordinate (anchor) at ($({#3})!{#5}!({#4})$);
            \node[darkmsgcircle, yshift=-5.0mm] at (anchor) {#6};
            \node[yshift=-2.0mm] at (anchor) {$\rightarrow$};
      }{}
      \ifthenelse{\isin{#1}{up} \AND \isin{#2}{right}}{
            \coordinate (anchor) at ($({#3})!{#5}!({#4})$);
            \node[darkmsgcircle, yshift=5.0mm] at (anchor) {#6};
            \node[yshift=2.0mm] at (anchor) {$\rightarrow$};
      }{}

      \ifthenelse{\isin{#1}{down} \AND \isin{#2}{left}}{
            \coordinate (anchor) at ($({#3})!{#5}!({#4})$);
            \node[darkmsgcircle, yshift=-5.0mm] at (anchor) {#6};
            \node[yshift=-2.0mm] at (anchor) {$\leftarrow$};
      }{}
      \ifthenelse{\isin{#1}{up} \AND \isin{#2}{left}}{
            \coordinate (anchor) at ($({#3})!{#5}!({#4})$);
            \node[darkmsgcircle, yshift=5.0mm] at (anchor) {#6};
            \node[yshift=2.0mm] at (anchor) {$\leftarrow$};
      }{}

      \ifthenelse{\isin{#1}{left} \AND \isin{#2}{up}}{
            \coordinate (anchor) at ($({#3})!{#5}!({#4})$);
            \node[darkmsgcircle, xshift=-5.0mm] at (anchor) {#6};
            \node[xshift=-1.5mm] at (anchor) {$\uparrow$};
      }{}
      \ifthenelse{\isin{#1}{right} \AND \isin{#2}{up}}{
            \coordinate (anchor) at ($({#3})!{#5}!({#4})$);
            \node[darkmsgcircle, xshift=5.0mm] at (anchor) {#6};
            \node[xshift=1.5mm] at (anchor) {$\uparrow$};
      }{}
}

\newcommand{\bwmsg}[6]{
      \ifthenelse{\isin{#1}{left} \AND \isin{#2}{down}}{
            \coordinate (anchor) at ($({#3})!{#5}!({#4})$);
            \node[msgdoublecircle, xshift=-5.5mm] at (anchor) {#6};
            \node[xshift=-1.5mm] at (anchor) {$\downarrow$};
      }{}
      \ifthenelse{\isin{#1}{right} \AND \isin{#2}{down}}{
            \coordinate (anchor) at ($({#3})!{#5}!({#4})$);
            \node[msgdoublecircle, xshift=5.5mm] at (anchor) {#6};
            \node[xshift=1.5mm] at (anchor) {$\downarrow$};
      }{}

      \ifthenelse{\isin{#1}{down} \AND \isin{#2}{right}}{
            \coordinate (anchor) at ($({#3})!{#5}!({#4})$);
            \node[msgdoublecircle, yshift=-6.0mm] at (anchor) {#6};
            \node[yshift=-2.0mm] at (anchor) {$\rightarrow$};
      }{}
      \ifthenelse{\isin{#1}{up} \AND \isin{#2}{right}}{
            \coordinate (anchor) at ($({#3})!{#5}!({#4})$);
            \node[msgdoublecircle, yshift=6.0mm] at (anchor) {#6};
            \node[yshift=2.0mm] at (anchor) {$\rightarrow$};
      }{}

      \ifthenelse{\isin{#1}{down} \AND \isin{#2}{left}}{
            \coordinate (anchor) at ($({#3})!{#5}!({#4})$);
            \node[msgdoublecircle, yshift=-6.0mm] at (anchor) {#6};
            \node[yshift=-2.0mm] at (anchor) {$\leftarrow$};
      }{}
      \ifthenelse{\isin{#1}{up} \AND \isin{#2}{left}}{
            \coordinate (anchor) at ($({#3})!{#5}!({#4})$);
            \node[msgdoublecircle, yshift=6.0mm] at (anchor) {#6};
            \node[yshift=2.0mm] at (anchor) {$\leftarrow$};
      }{}

      \ifthenelse{\isin{#1}{left} \AND \isin{#2}{up}}{
            \coordinate (anchor) at ($({#3})!{#5}!({#4})$);
            \node[msgdoublecircle, xshift=-5.5mm] at (anchor) {#6};
            \node[xshift=-1.5mm] at (anchor) {$\uparrow$};
      }{}
      \ifthenelse{\isin{#1}{right} \AND \isin{#2}{up}}{
            \coordinate (anchor) at ($({#3})!{#5}!({#4})$);
            \node[msgdoublecircle, xshift=5.5mm] at (anchor) {#6};
            \node[xshift=1.5mm] at (anchor) {$\uparrow$};
      }{}
}

\newcommand{\bwdarkmsg}[6]{
      \ifthenelse{\isin{#1}{left} \AND \isin{#2}{down}}{
            \coordinate (anchor) at ($({#3})!{#5}!({#4})$);
            \node[darkmsgdoublecircle, xshift=-5.5mm] at (anchor) {#6};
            \node[xshift=-1.5mm] at (anchor) {$\downarrow$};
      }{}
      \ifthenelse{\isin{#1}{right} \AND \isin{#2}{down}}{
            \coordinate (anchor) at ($({#3})!{#5}!({#4})$);
            \node[darkmsgdoublecircle, xshift=5.5mm] at (anchor) {#6};
            \node[xshift=1.5mm] at (anchor) {$\downarrow$};
      }{}

      \ifthenelse{\isin{#1}{down} \AND \isin{#2}{right}}{
            \coordinate (anchor) at ($({#3})!{#5}!({#4})$);
            \node[darkmsgdoublecircle, yshift=-6.0mm] at (anchor) {#6};
            \node[yshift=-2.0mm] at (anchor) {$\rightarrow$};
      }{}
      \ifthenelse{\isin{#1}{up} \AND \isin{#2}{right}}{
            \coordinate (anchor) at ($({#3})!{#5}!({#4})$);
            \node[darkmsgdoublecircle, yshift=6.0mm] at (anchor) {#6};
            \node[yshift=2.0mm] at (anchor) {$\rightarrow$};
      }{}

      \ifthenelse{\isin{#1}{down} \AND \isin{#2}{left}}{
            \coordinate (anchor) at ($({#3})!{#5}!({#4})$);
            \node[darkmsgdoublecircle, yshift=-6.0mm] at (anchor) {#6};
            \node[yshift=-2.0mm] at (anchor) {$\leftarrow$};
      }{}
      \ifthenelse{\isin{#1}{up} \AND \isin{#2}{left}}{
            \coordinate (anchor) at ($({#3})!{#5}!({#4})$);
            \node[darkmsgdoublecircle, yshift=6.0mm] at (anchor) {#6};
            \node[yshift=2.0mm] at (anchor) {$\leftarrow$};
      }{}

      \ifthenelse{\isin{#1}{left} \AND \isin{#2}{up}}{
            \coordinate (anchor) at ($({#3})!{#5}!({#4})$);
            \node[darkmsgdoublecircle, xshift=-5.5mm] at (anchor) {#6};
            \node[xshift=-1.5mm] at (anchor) {$\uparrow$};
      }{}
      \ifthenelse{\isin{#1}{right} \AND \isin{#2}{up}}{
            \coordinate (anchor) at ($({#3})!{#5}!({#4})$);
            \node[darkmsgdoublecircle, xshift=5.5mm] at (anchor) {#6};
            \node[xshift=1.5mm] at (anchor) {$\uparrow$};
      }{}
}

\makeatletter
\DeclareRobustCommand{\cev}[1]{%
  {\mathpalette\do@cev{#1}}%
}
\newcommand{\do@cev}[2]{%
  \vbox{\offinterlineskip
    \sbox\z@{$\m@th#1 x$}%
    \ialign{##\cr
      \hidewidth\reflectbox{$\m@th#1\vec{}\mkern4mu$}\hidewidth\cr
      \noalign{\kern-\ht\z@}
      $\m@th#1#2$\cr
    }%
  }%
}

\newcommand{\midi}[0]{%
    {\,|\,}%
}

\title{Principled Pruning of Bayesian Neural Networks through Variational Free Energy Minimization}

\date{\today}	% Here you can change the date presented in the paper title
\author{
  Jim Beckers$^{1}$ \qquad Bart van Erp$^{*,1}$ \qquad Ziyue Zhao$^{2}$ \qquad Kirill Kondrashov$^{2}$ \qquad Bert de Vries$^{1, 2}$ \\
  $^{1}$Department of Electrical Engineering, Eindhoven University of Technology, Eindhoven, The Netherlands \\
  $^{2}$GN Hearing, Eindhoven, The Netherlands
  \thanks{Correspondence: \href{mailto:b.v.erp@tue.nl}{b.v.erp@tue.nl}}
}

\renewcommand{\shorttitle}{Principled Pruning of Bayesian Neural Networks}

\begin{document}
\maketitle

% Added to get the footnotes in the text to start at 1
\setcounter{footnote}{0}

\begin{abstract}
    Bayesian model reduction provides an efficient approach for comparing the performance of all nested sub-models of a model, without re-evaluating any of these sub-models.
    Until now, Bayesian model reduction has been applied mainly in the computational neuroscience community on simple models.
    In this paper, we formulate and apply Bayesian model reduction to perform principled pruning of Bayesian neural networks, based on variational free energy minimization.
    Direct application of Bayesian model reduction, however, gives rise to approximation errors.
    Therefore, a novel iterative pruning algorithm is presented to alleviate the problems arising with naive Bayesian model reduction, as supported experimentally on the publicly available UCI datasets for different inference algorithms.
    This novel parameter pruning scheme solves the shortcomings of current state-of-the-art pruning methods that are used by the signal processing community.
    The proposed approach has a clear stopping criterion and minimizes the same objective that is used during training.
    Next to these benefits, our experiments indicate better model performance in comparison to state-of-the-art pruning schemes.
\end{abstract}

\ifIEEE
\else
\fi
\ifIEEE
    \section{INTRODUCTION}\label{sec:introduction}
\else
    \section{Introduction}\label{sec:introduction}
\fi

\ifIEEE
    \IEEEPARstart{C}{urrent}
\else
    Current
\fi
% application
state-of-the-art neural networks excel in a variety of signal processing tasks, such as audio processing \cite{radford_robust_2023}.
Yet, the large size of these state-of-the-art models limits their direct applicability in storage- and power-constrained devices, such as hearing aids.
For example, neural networks for hearing aids are often relatively small \cite{fedorov_tinylstms_2020, diehl_restoring_2023, westhausen_low_2023}, where performance is sacrificed for real-time and efficient operations.
These smaller architectures limit expressivity and assume that all parameters are used efficiently.
Another strategy is to start with a large network and to compress the network in order to utilize less storage and power resources \cite{cheng_survey_2020, mishra_survey_2020, zhang_compacting_2021}.
One approach to network compression, parameter pruning, removes parameters from the network in order to lower computational and storage demands, whilst aiming to retain good-enough performance.
Unfortunately, these pruning methods for neural networks are often based on heuristic criteria with tunable thresholds, thus turning parameter pruning into a lengthy (and costly) trial-and-error procedure.

% pruning in BNNs
Bayesian neural networks (BNNs) are the Bayesian equivalent of neural networks, in which the beliefs about the parameter values of the network are represented by probability distributions.
BNNs enjoy a proper model performance criterion: the variational free energy (VFE).
VFE provides a yardstick to compare different models regarding their performance on ``accuracy minus complexity".
In other words, optimizing VFE leads to models that maximize data fit (accuracy) and simultaneously minimize model complexity (Occam's razor is built-in).
These objectives align with good modeling practices.
Consequently, BNNs provide several advantages over regular neural networks such as allowing for incorporating prior knowledge, robustness to overfitting and straightforward online or continual learning \cite{jospin_hands-bayesian_2021}.
Most importantly, they allow for more principled compression methods as they retain beliefs about their parameter values.
As of today, pruning in BNNs has seen little attention when compared to pruning in regular neural networks \cite{louizos_bayesian_2017, chirkova_bayesian_2018}, as it is still an active area of research \cite{wu_deterministic_2019, haussmann_sampling-free_2019}.
Current state-of-the-art parameter pruning methods in BNNs lack some vital characteristics. 
Firstly, the proposed pruning objectives do not coincide with the objectives used during the training of the BNN \cite{graves_practical_2011, blundell_weight_2015, nalisnick_priors_2018}.
Additionally, current pruning strategies do not provide a clear optimal pruning rate as they are based on heuristic criteria with tunable thresholds.

% Bayesian model reduction
Originating from the computational neuroscience literature, \emph{Bayesian model reduction} (BMR) \cite{friston_post_2011, friston_bayesian_2018} provides an approximate method for comparing VFE among all nested sub-models of a model, without re-evaluating any of the sub-models.
In short, we only train a model and BMR provides a cheap and principled method to select pruned sub-models with better VFE than the large model.
BMR enjoys biological plausibility as described in various papers in the neuroscience literature \cite{friston_post_2011, friston_bayesian_2018}.
So far, BMR has received little attention outside of the neuroscience literature and has therefore not been applied to large complex models.
If properly developed for BNNs, it would also provide a strong method for fast and principled pruning of trained BNNs to smaller and better performing BNNs, to enable adoption of neural networks in resource-constrained devices.

% contributions
This paper extends Bayesian model reduction to parameter pruning in BNNs that tackles the above problems with current state-of-the-art pruning approaches.
Specifically, we use BMR to evaluate the performance of pruned BNNs by shrinking the prior distributions over their parameters.
Direct application of BMR exposes fundamental limitations when dealing with factorized posterior distributions, as we will show in Section~\ref{sec:pruning}.
This paper alleviates this problem by proposing a novel pruning procedure for BNNs, where training and pruning share the same objective.
We experimentally show that our principled pruning procedure for BNNs achieves the lowest VFE in comparison to conventional pruning heuristics on the publicly available UCI datasets~\cite{Dua:2019}.
Furthermore we show that our procedure is invariant to different inference algorithms.
Before introducing our principled pruning scheme in Section~\ref{sec:pruning}, we first specify the model of the BNN in Section~\ref{sec:model} and touch upon the inference procedure in Section~\ref{sec:inference}.
We postpone an overview of the related work to Section~\ref{sec:related} since we believe that the reader will appreciate the comparison to existing literature more after the introduction of our proposed pruning algorithm.
We verify and validate our contributions in Section~\ref{sec:experiments}.
Section~\ref{sec:discussion} discusses our approach and provides opportunities for future research.
Section~\ref{sec:conclusion} concludes the paper.
\ifIEEE
    \section{MODEL SPECIFICATION}\label{sec:model}
\else
    \section{Model specification}\label{sec:model}
\fi
Before the presentation of our novel parameter pruning scheme in Section~\ref{sec:pruning}, we will first formally specify the probabilistic model of our BNN in this section.
Then in Section~\ref{sec:inference}, the probabilistic inference procedure used to perform computations in this model is elaborated upon.

Consider the likelihood function of the BNN
\begin{equation} \label{eq:likelihood}
    p(Y \midi \bm{\theta}, \bm{\gamma}) = \prod_{n=1}^N \mathcal{N} \left( \bm{y}_n \midi f_{\bm{\theta}}(\bm{x}_n), \Lambda^{-1} \right),
\end{equation}
where $\mathcal{N}(\cdot \midi \mu, \Sigma)$ depicts a normal distribution with mean $\mu$ and covariance $\Sigma$.
$X = \{\bm{x}_1, \bm{x}_2, \ldots, \bm{x}_N\}$ and $Y = \{\bm{y}_1, \bm{y}_2, \ldots, \bm{y}_N\}$ denote the sets of possibly multivariate inputs $\bm{x}_n\in\mathbb{R}^D$ and outputs $\bm{y}_n\in\mathbb{R}^M$ of the BNN, respectively.
Here, $D$ and $M$ refer to the dimensionality of the input and output samples, and $N$ represents the number of data samples.
The precision matrix $\Lambda$ represents a diagonal matrix with positive elements $\bm{\gamma} = [\gamma_1, \gamma_2, \ldots, \gamma_M]^\top \in \mathbb{R}^M_{>0}$.
The most important term $f_{\bm{\theta}}(\cdot)$ represents the underlying neural network parameterized by the set of model parameters $\bm{\theta}$.
The pruning algorithm as presented in Section~\ref{sec:pruning} is model agnostic and therefore the exact specification of $f_{\bm{\theta}}(\cdot)$ will only be presented in Section~\ref{sec:experiments:setup} as the architecture differs between experiments.
Although \eqref{eq:likelihood} solely refers to regression models, extensions to models suitable for classification are straightforward and do not change the specification of the pruning algorithm of Section~\ref{sec:pruning}.

In order to complete the specification of the BNN, we will define prior distributions over $\bm{\theta}$ and $\bm{\gamma}$.
We assume the factorized prior over the model parameters
\begin{equation} \label{eq:prior-params}
    p(\bm{\theta}) = \prod_{l=1}^L \prod_{\theta \in \bm{\theta}^{(l)}} \mathcal{N}\left(\theta \midi \mu_\theta, \sigma^2_\theta\right),
\end{equation}
which factorizes over the $L$ layers in the model and over the individual parameters within each layer $\theta\in\bm{\theta}^{(l)}$.
Additionally, we set the prior distribution
\begin{equation} \label{eq:prior-precision}
    p(\bm{\gamma}) = \prod_{m=1}^{M} \Gamma (\gamma_m \midi \alpha_m, \beta_m),
\end{equation}
where $\Gamma(\cdot \midi \alpha, \beta)$ represents a gamma distribution with shape and rate parameters $\alpha$ and $\beta$, respectively.
\ifIEEE
    \section{PROBABILISTIC INFERENCE}\label{sec:inference}
\else
    \section{Probabilistic inference}\label{sec:inference}
\fi
\bart{tentative: shorten section (less textbook-ish)}
Probabilistic inference concerns the computation of posterior and predictive distributions in the BNN specified by \eqref{eq:likelihood}-\eqref{eq:prior-precision}.
The exact posterior distribution $p(\bm{\theta}, \bm{\gamma} \midi Y)$ can be computed using Bayes' rule as
\begin{equation}
    p(\bm{\theta}, \bm{\gamma} \midi Y) = \frac{p(Y \midi \bm{\theta}, \bm{\gamma}) \, p(\bm{\theta}) \, p(\bm{\gamma})}{p(Y)}.
\end{equation}
However, the computation of this posterior distribution is intractable for most choices of $f_{\bm{\theta}}(\cdot)$.
A solution to this problem is obtained by approximating the exact posterior distribution $p(\bm{\theta}, \bm{\gamma} \midi Y)$ with a variational posterior $q(\bm{\theta}, \bm{\gamma}) \in \mathcal{Q}$, which is constrained to a simplified family of distributions $\mathcal{Q}$.
We find the optimal solution for $q(\bm{\theta}, \bm{\gamma})$ by minimizing the variational free energy (VFE), given by
\begin{equation} \label{eq:VFE}
    \mathrm{F}[p, q] = \int q(\bm{\theta}, \bm{\gamma}) \ln \frac{q(\bm{\theta},\bm{\gamma})}{p(Y\midi \bm{\theta},\bm{\gamma})\ p(\bm{\theta})\, p(\bm{\gamma})}\, \mathrm{d}\bm{\theta}\, \mathrm{d}\bm{\gamma}.
\end{equation}

Note that the VFE can be decomposed into
\begin{equation} \label{eq:VFE-posterior}
    \mathrm{F}[p,q] = \underbrace{\mathrm{KL}\left[ q(\bm{\theta}, \bm{\gamma}) \,\|\, p(\bm{\theta}, \bm{\gamma} \midi Y) \right]}_{\text{posterior divergence}} - \underbrace{\ln p(Y)}_{\text{log-evidence}},
\end{equation}
which, since the Kullback-Leibler ($\mathrm{KL}$) divergence is always non-negative, illustrates that the VFE forms an upper bound on the negative log-evidence.
Therefore the VFE is equivalent to the well-known negative evidence lower bound (ELBO) criterion in the machine learning community.
Minimization of the VFE minimizes the $\mathrm{KL}$-divergence between the variational posterior $q(\bm{\theta}, \bm{\gamma})$ and the exact posterior distribution $p(\bm{\theta}, \bm{\gamma} \midi Y)$, and reduces the gap between the VFE and the negative log-evidence, which is determined by the observations and the model specification.

Optimization of \eqref{eq:VFE-posterior} is not possibe as the exact posterior $p(\bm{\theta}, \bm{\gamma} \midi Y)$ is unknown.
Therefore we proceed to derive a minimization procedure through the decomposition 
\ifIEEE
    \begin{multline}\label{eq:VFE-complexity}
        \mathrm{F}[p,q] = \underbrace{\mathrm{KL} \left[q(\bm{\theta},\bm{\gamma}) \,\|\, p(\bm{\theta})p(\bm{\gamma}) \right] }_{\text{complexity}} \\
        - \underbrace{\mathbb{E}_{q(\bm{\theta})q(\bm{\gamma})} \left[ \ln p(Y \midi \bm{\theta}, \bm{\gamma}) \right]}_{\text{accuracy}}.
    \end{multline}
\else
    \begin{equation} \label{eq:VFE-complexity}
        \mathrm{F}[p,q] = \underbrace{\mathrm{KL} \left[q(\bm{\theta},\bm{\gamma}) \,\|\, p(\bm{\theta})p(\bm{\gamma}) \right] }_{\text{complexity}} 
        - \underbrace{\mathbb{E}_{q(\bm{\theta})q(\bm{\gamma})} \left[ \ln p(Y \midi \bm{\theta}, \bm{\gamma}) \right]}_{\text{accuracy}}.
    \end{equation}
\fi
Furthermore we assume that the variational posterior has a mean-field factorization $q(\bm{\theta}, \bm{\gamma}) = q(\bm{\theta}) q(\bm{\gamma})$, where the individual factors are further factorized as 
\begin{subequations}
    \begin{equation} \label{eq:q:params}
        q(\bm{\theta}) = \prod_{l=1}^L \prod_{\theta \in \bm{\theta}^{(l)}} \mathcal{N}(\theta \midi \hat{\mu}_\theta, \hat{\sigma}^2_\theta),
    \end{equation}
    \begin{equation}
        q(\bm{\gamma}) = \prod_{m=1}^M \Gamma(\gamma_m \midi \hat{\alpha}_m, \hat{\beta}_m).
    \end{equation}
\end{subequations}
Here the $\hat{\cdot}$ accent distinguishes the parameters of the variational posterior distributions from the parameters of the prior distributions.
Under these simplifying assumptions the $\mathrm{KL}$-divergence between the variational posterior and prior distribution has a closed-form solution for the chosen family of distributions.
The accuracy term can be approximated in different ways, for example using variance backpropagation \cite{haussmann_sampling-free_2019}, where the first and second central moments are propagated through the neural network, or Bayes-by-backprop \cite{blundell_weight_2015}, which employs sampling for computing the output.
Using these approximations, the VFE can be optimized over the parameters of the variational posterior distributions using (stochastic) gradient descent, also known as stochastic variational inference \cite{hoffman_stochastic_2013, blei_variational_2017-1}.
\ifIEEE
    \section{PARAMETER PRUNING}\label{sec:pruning}
\else
    \section{Parameter pruning}\label{sec:pruning}
\fi
Now that we have specified the probabilistic model of the BNN in Section~\ref{sec:model} and have elaborated upon the training procedure in Section~\ref{sec:inference}, we can introduce our novel parameter pruning scheme.
This parameter pruning scheme shares its objective with the training stage and has a clear stopping criterion, solving the issues with conventional pruning approaches as highlighted in Section~\ref{sec:introduction}.

\begin{figure*}[t]
    \centering
    \resizebox{\textwidth}{!}{\input{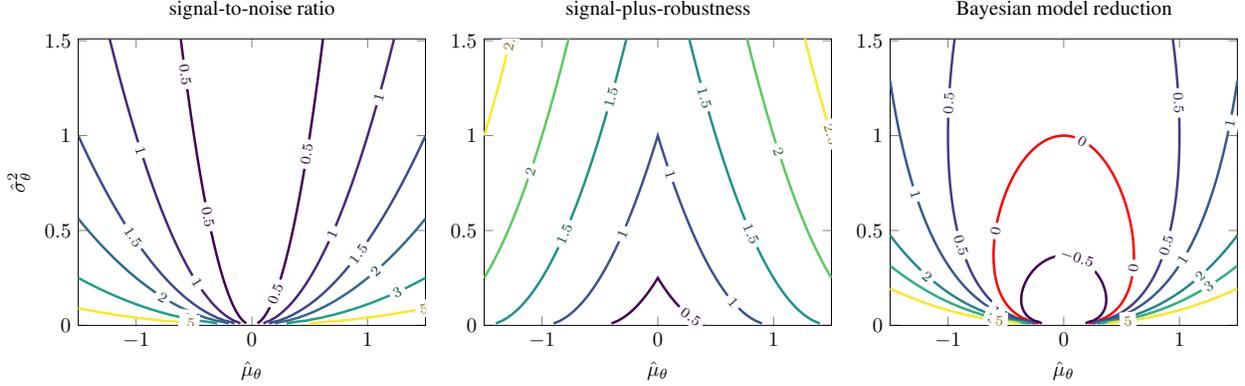}}
    \vspace*{-4mm}\caption{An overview of the pruning objectives signal-to-noise ratio, signal-plus-robustness and Bayesian model reduction. The prior distribution was set to $p(\theta) = \mathcal{N}(\theta \midi 0, 1)$ and the new prior to $\tilde{p}(\theta) = \mathcal{N}(\theta\midi 0, \varepsilon)$, with $\varepsilon=10^{-16}$. The pruning objectives have been computed with respect to the variational posterior distribution $q(\theta) = \mathcal{N}(\theta\midi \hat{\mu}_\theta, \hat{\sigma}^2_\theta)$. All three objectives prune from low to high values. Only Bayesian model reduction has a clear stopping criterion, located at zero, as indicated by the red contour in the right plot. The interior of this contour is subject to pruning.}
    \label{fig:pruning-comparison}
\end{figure*}

Our pruning scheme is based on Bayesian model reduction \cite{friston_post_2011, friston_bayesian_2018}.
After training the full model, BMR supports efficient recomputation of posterior distributions and model evidence upon changes to the model prior.
Consider the likelihood function $p(Y \midi \bm{\theta})$ in \eqref{eq:likelihood}, where the $\bm{\gamma}$ argument is omitted to simplify notation, without removing it from the model.
We omit $\bm{\gamma}$ to focus on $\bm{\theta}$, which encompasses the vast majority of model parameters that we wish to prune.
Based on the prior distribution $p(\bm{\theta})$, we can compute the posterior $p(\bm{\theta} \midi Y)$ and model evidence $p(Y)$ using Bayes' rule.
Let us now introduce a new or reduced prior $\tilde{p}(\bm{\theta})$, where the $\tilde{\cdot}$ accent refers to the new situation.
This prior shares the same likelihood function as before, meaning that $p(Y\midi \bm{\theta}) = \tilde{p}(Y \midi \bm{\theta})$ holds.
Instead of computing the new posterior $\tilde{p}(\bm{\theta}\midi Y)$ and new evidence $\tilde{p}(Y)$ from scratch again, which is often a costly operation, BMR uses the original posterior and evidence in order to compute the new posterior and evidence in a simplified manner.
The shared likelihood function allows us to compute the new posterior and log-evidence as \cite{friston_post_2011, friston_bayesian_2018}
\begin{subequations} \label{eq:bmr-p}
    \begin{equation}
        \tilde{p}(\bm{\theta} \midi Y) = p(\bm{\theta} \midi Y) \frac{\tilde{p}(\bm{\theta})}{p(\bm{\theta})} \frac{p(Y)}{\tilde{p}(Y)},
    \end{equation}
    \begin{equation} \label{eq:bmr:p-evidence}
        \ln \tilde{p}(Y) = \ln p(Y) + \ln \int p(\bm{\theta} \midi Y) \frac{\tilde{p}(\bm{\theta})}{p(\bm{\theta})} \mathrm{d} \bm{\theta}.
    \end{equation}
\end{subequations}
Detailed derivations are presented in Appendix~\ref{app:bmr}.

\begin{algorithm*}[t]
    \setstretch{1.15}
    \caption{Iterative training and pruning of a Bayesian neural network by VFE minimization.} \label{alg:bmr}
    \begin{algorithmic}[1]
        \Inputs{Bayesian neural network model with parameter prior $p(\bm{\theta})$}
        \While {Parameters were pruned in previous iteration}
            \State Train (pruned) Bayesian neural network
            \For{$\theta \in \bm{\theta}$} \Comment{Loop over active parameters}
                \State $\tilde{p}_\theta(\bm{\theta}) \gets p(\bm{\theta} \backslash \theta) \tilde{p}(\theta)$ \Comment{Specify new prior}
                \State Compute $\Delta\mathrm{F}_\theta$ for $\tilde{p}_\theta(\bm{\theta})$ using \eqref{eq:bmr-q-vfe}
                \If{$\Delta\mathrm{F}_\theta \leq 0$} 
                    \State $\bm{\theta} \gets \bm{\theta} \backslash \theta$ \Comment{Prune parameter}
                \EndIf
            \EndFor
        \EndWhile
    \end{algorithmic}
\end{algorithm*}

In BNNs we generally do not have access to the intractable $p(\bm{\theta} \midi Y)$ and $p(Y)$ as explained in Section~\ref{sec:inference}.
Therefore we will need to approximate the BMR procedure by substituting the variational posterior $q(\bm{\theta})$ for the exact posterior $p(\bm{\theta} \midi Y)$ and the VFE $\mathrm{F}[p,q]$ for the negative log-evidence $-\ln p(Y)$, based on the bound in \eqref{eq:VFE-posterior}.
Resulting from these substitutions, we obtain expressions for the new variational posterior $\tilde{q}(\bm{\theta})$ and for the change in VFE $\Delta\mathrm{F}$ as
\begin{subequations} \label{eq:bmr-q}
    \begin{equation} \label{eq:bmr-q-posterior}
        \tilde{q}(\bm{\theta}) \approx q(\bm{\theta}) \frac{\tilde{p}(\bm{\theta})}{p(\bm{\theta})} \exp\{\Delta \mathrm{F}\},
    \end{equation}
    \begin{equation} \label{eq:bmr-q-vfe}
        \Delta\mathrm{F} = \mathrm{F}[p,q] - \mathrm{F}[\tilde{p}, \tilde{q}] \approx - \ln \int q(\bm{\theta}) \frac{\tilde{p}(\bm{\theta})}{p(\bm{\theta})} \mathrm{d}\bm{\theta}.
    \end{equation}
\end{subequations}
Intuitively $\exp\{\Delta \mathrm{F}\}$ can be obtained from \eqref{eq:bmr-q-posterior} using the normalization property of $\tilde{q}(\bm{\theta})$.
As a result of the factorization assumptions in \eqref{eq:prior-params} and \eqref{eq:q:params}, the change in VFE decomposes in $\Delta \mathrm{F} \approx \sum_{\theta\in\bm{\theta}} \Delta \mathrm{F}_\theta$, where $\Delta \mathrm{F}_\theta$ is the change in VFE obtained by pruning a single parameter $\theta$.

Parameter pruning is performed by evaluating $\Delta\mathrm{F}_\theta$ for each parameter individually, where all parameters with $\Delta\mathrm{F}_\theta \leq 0$ are pruned.
Despite the seeming complexity of \eqref{eq:bmr-q}, the expressions submit to simple closed-form solutions when the prior and variational posterior distributions are assumed to be in the same family of distributions as presented in \cite[Table 1]{friston_bayesian_2018}.
Appendix~\ref{app:bmr-gaussian} presents the closed-form solutions for normal distributions.
For our choice of normal prior and variational posterior distributions, we perform pruning by setting the individual prior distributions to $\tilde{p}(\theta) = \mathcal{N}(\theta \midi 0, \varepsilon)$, where $\theta \in \bm{\theta}$ is a single parameter in the model and where $\varepsilon$ represents a sufficiently small number close to zero.
Basically, the new prior is set to cover a small region around zero, preventing the corresponding variational posterior to differ significantly upon new observations.
For each of the parameters $\theta \in \bm{\theta}$ the difference in VFE $\Delta\mathrm{F}_\theta$ is evaluated.
Pruning starts from the lowest values of $\Delta\mathrm{F}_\theta$ and stops when $\Delta\mathrm{F}_\theta$ surpasses zero because then an improvement in VFE is no longer possible.
Figure~\ref{fig:pruning-comparison} shows an intuitive comparison between BMR in \eqref{eq:bmr-q} and current state-of-the-art pruning methods for common choices of prior distributions.
Specifically, we compare against the signal-to-noise ratio (SNR) \cite{graves_practical_2011, blundell_weight_2015} and signal-plus-robustness (SPR) metric \cite[Ch. 4]{nalisnick_priors_2018}.
More information about these ranking functions is presented in Section~\ref{sec:related}.

The approximation in \eqref{eq:bmr-q}, resulting from the unavailability of the exact posterior distribution, leads to estimation discrepancies for the VFE difference.
For BNNs, the accumulating effects of this approximation lead to a divergence between the approximated and actual difference inVFE.
Simultaneously, because the probabilistic model is changed during pruning, the obtained variational posterior $q(\bm{\theta})$ no longer is guaranteed to be located in a stationary point of the VFE, which was obtained after training.

In order to cope with these problems, we propose an update to BMR by using an iterative process where we do not use BMR solely as a post-hoc method but incorporate it into the training stage of BNNs, similarly as in \cite{graves_practical_2011, riera_dnn_2022}.
The process starts by training the BNN until convergence of the VFE.
It then computes the VFE changes $\Delta\mathrm{F}_\theta$ for all $\theta\in\bm{\theta}$ individually and greedily prunes the parameters where $\Delta \mathrm{F}_\theta \leq 0$.
The process of training and pruning is repeated until no more parameters can be pruned.
This iterative procedure is formalized in Algorithm~\ref{alg:bmr}.

% As stated in line 4 of the Algorithm, we loop over all active parameters iteratively. 
% Since our initial assumption was that all model parameters are independent, we can process them sequentially regardless of the ordering of the parameters. 
% In addition, the change in variational free energy can be computed per individual model parameter, independently of the other model parameters~\cite{friston_post_2011}. 
% We do note that the algorithm is not guaranteed to find the global optimum, but rather a local optimum in an efficient manner, as we do not perform an exhaustive search over the complete model space.
\ifIEEE
    \section{EXPERIMENTS}\label{sec:experiments}
\else
    \section{Experiments}\label{sec:experiments}
\fi
This section will present experimental results for the verification of our new pruning algorithm (Subsection~\ref{sec:experiments:divergence}), for the robustness and performance of our approach using different inference algorithms (Subsection~\ref{sec:experiments:inference}) and for the comparison to state-of-the-art methods, such as SNR and SPR (Subsection~\ref{sec:experiments:comparison}).

\ifIEEE
    \subsection{EXPERIMENTAL SETUP}\label{sec:experiments:setup}
\else
    \subsection{Experimental setup}\label{sec:experiments:setup}
\fi
All experiments were performed in \texttt{Python}~(v.3.9) using \texttt{TensorFlow}~(v.2.9)\footnote{All experiments are publicly available at \href{https://github.com/biaslab/PrincipledPruningBNN}{https://github.com/biaslab/PrincipledPruningBNN}.}.
The network $f_{\bm{\theta}}(\cdot)$ in \eqref{eq:likelihood} is represented either by fully connected networks (FCNs) or by recurrent neural networks (RNNs).
We followed the setup of~\cite{hernandez-lobato_probabilistic_2015}, which uses a subset of the publicly available UCI datasets~\cite{Dua:2019} and an FCN with a single hidden layer of 50 units.
For the RNN experiments, we used a synthetic dataset consisting of time-periodic signals and a neural network with a single Gated Recurrent Unit (GRU)~\cite{cho_learning_2014} layer of 16 units.
We used a sequence length of 8 to predict the next single sample.
Both models were trained by minimizing \eqref{eq:VFE-complexity} with Adam \cite{Kingma2015AdamAM} using variance backpropagation \cite{haussmann_sampling-free_2019} and Bayes-by-backprop \cite{blundell_weight_2015} with both global and local parameterizations \cite{kingma_auto-encoding_2014, kingma_variational_2015}.
For computing the accuracy term in \eqref{eq:VFE-complexity} with Bayes-by-backprop, we use a single sample during training and 10 samples during the performance evaluation.

\begin{figure*}[t]
    \centering
    \resizebox{\textwidth}{!}{\input{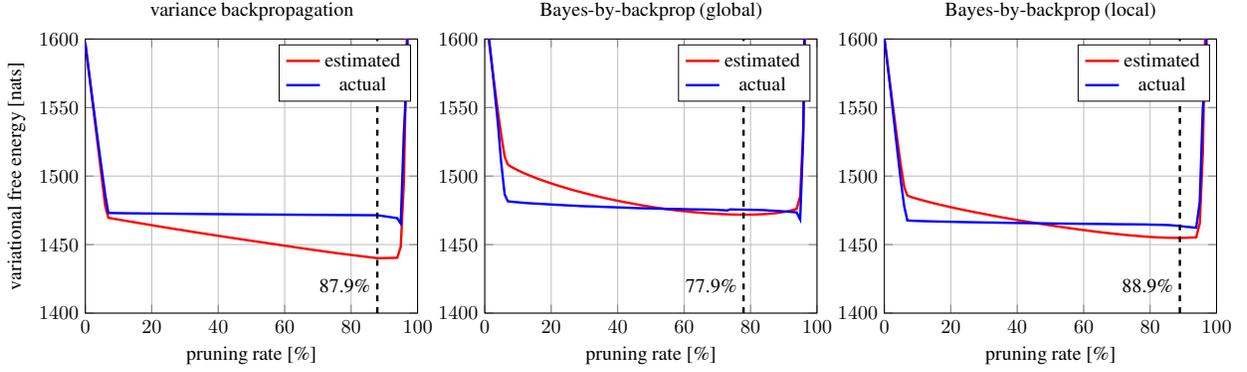}}
    \vspace*{-4mm}\caption{Estimated and actual variational free energy for different pruning rates for the \texttt{boston} dataset. The different plots illustrate the different inference algorithms: variance backpropagation \cite{haussmann_sampling-free_2019} and Bayes-by-backprop \cite{blundell_weight_2015} with global \cite{kingma_auto-encoding_2014} and local \cite{kingma_variational_2015} reparameterization. The dashed line denotes the stopping criterion for a single pruning iteration. In all plots, the estimated and actual optimal pruning rates differ, establishing the need of Algorithm~\ref{alg:bmr}.}
    \label{fig:experiments:divergence}
\end{figure*}

\ifIEEE
    \subsection{DIVERGENCE ASSESSMENT} \label{sec:experiments:divergence}
\else
    \subsection{Divergence assessment} \label{sec:experiments:divergence}
\fi
The approximation in and recursive usage of \eqref{eq:bmr-q} leads to a divergence between the approximated and actual VFE for different pruning rates.
To illustrate this divergence, we perform parameter pruning with increments of 1\% and compute the corresponding estimated and exact VFE.
Figure~\ref{fig:experiments:divergence} shows the discrepancy between the estimated and actual VFE for different pruning rates on the \texttt{boston} dataset for the different inference algorithms.
This discrepancy can be observed for all datasets.
As a result of this discrepancy, a single round of pruning until the optimal stopping criterion has been reached, might not yield the minimum of the VFE.
This establishes the need of the iterative Algorithm~\ref{alg:bmr}.

\ifIEEE
    \subsection{ROBUSTNESS AND PERFORMANCE} \label{sec:experiments:inference}
\else
    \subsection{Robustness and performance} \label{sec:experiments:inference}
\fi
Table~\ref{tab:experiments:performance} shows the performance of our proposed pruning algorithm for different inference algorithms and datasets.
We perform the pruning for a single iteration of the BMR algorithm according to Algorithm~\ref{alg:bmr}, until the stopping criterion has been reached.
For all datasets, we achieve a lower VFE with respect to the full model already for a single iteration of the BMR algorithm.
Our proposed iterative procedure performs better in all cases, achieving both lower VFE values as well as higher pruning rates.
These results are consistent over the different datasets and different inference algorithms, underlining the robustness of the proposed approach.
Oftentimes the first iteration has the biggest contribution, however, this depends significantly on the dataset and model.
Depending on the application one could choose to stop the pruning algorithm earlier.

{\setlength{\tabcolsep}{2.5pt}      %
 \renewcommand{\arraystretch}{1.1}  %
\begin{table*}[t]
    \centering
    {\setstretch{0.7} %
    \caption{Performance overview of a single iteration of the Bayesian model reduction algorithm and of the iterative Algorithm~\ref{alg:bmr} across different datasets \cite{Dua:2019} and inference algorithms. Performance is indicated by the obtained variational free energy and pruning rate (within brackets) after pruning. Variational free energy values after regular training are listed as reference.}\label{tab:experiments:performance}}
    \resizebox{\textwidth}{!}{%
    \begin{tabular}{l r r r r r r r r r}
        \hline
         & \multicolumn{3}{c}{\textit{variance backpropagation}} & \multicolumn{3}{c}{\textit{Bayes-by-backprop (global)}} & \multicolumn{3}{c}{\textit{Bayes-by-backprop (local)}} \\
        \cmidrule(rl){2-4} \cmidrule(rl){5-7} \cmidrule(rl){8-10}
        \textbf{dataset} & \multicolumn{1}{c}{\textbf{start}} & \multicolumn{1}{c}{\textbf{1 iteration}} & \multicolumn{1}{c}{\textbf{iterative}} & \multicolumn{1}{c}{\textbf{start}} & \multicolumn{1}{c}{\textbf{1 iteration}} & \multicolumn{1}{c}{\textbf{iterative}} & \multicolumn{1}{c}{\textbf{start}} & \multicolumn{1}{c}{\textbf{1 iteration}} & \multicolumn{1}{c}{\textbf{iterative}} \\
        \hline\hline \\ [-1.1em]
        \texttt{boston}     & 1.597 & 1.471 (88\%) & 1.454 (94\%)           & 1.624 & 1.475 (78\%) & 1.456 (93\%)       & 1.601 & 1.464 (89\%) & 1.451 (94\%)       \\
        \texttt{concrete}   & 3.707 & 3.626 (92\%) & 3.558 (93\%)           & 3.716 & 3.623 (79\%) & 3.540 (92\%)       & 3.716 & 3.673 (84\%) & 3.555 (92\%)       \\
        \texttt{energy}     & 4.581 & 4.217 (75\%) & 4.026 (83\%)           & 4.208 & 3.827 (75\%) & 3.392 (85\%)       & 4.255 & 3.870 (75\%) & 3.414 (84\%)       \\
        \texttt{kin8nm}     & -5.906 & -6.303 (71\%) & -6.377 (80\%)        & -5.286 & -5.681 (75\%) & -5.894 (85\%)    & -5.419 & -5.822 (74\%) & -5.988 (84\%)    \\
        \texttt{naval}      & -141.498 & -142.082 (47\%) & -142.870 (98\%)     & -139.557 & -141.783 (81\%) & -142.902 (99\%)      & -140.424 & -142.042 (60\%) & -142.875 (99\%)      \\
        \texttt{powerplant} & 29.267 & 29.164 (17\%) & 28.517 (53\%)        & 28.895 & 28.771 (28\%) & 28.190 (68\%)    & 29.251 & 29.186 (10\%) & 28.262 (74\%)    \\
        \texttt{wine}       & 1.902 & 1.679 (90\%) & 1.633 (98\%)           & 2.171 & 1.695 (88\%) & 1.578 (98\%)       & 1.995 & 1.648 (89\%) & 1.581 (98\%)       \\
        \texttt{yacht}      & 882 & 656 (78\%) & 572 (89\%)                 & 829 & 632 (80\%) & 625 (93\%)             & 832 & 632 (77\%) & 547 (89\%)             \\
        [-1.2em] \\ \hdashline \\ [-1.1em]
        \texttt{sine}       & -1.081 & 39.579 (89\%) & -1.221 (93\%)        & -943 & -1.125 (65\%) & -1.194 (89\%)      & -1.053 & -1.191 (76\%) & -1.225 (89\%)    \\
        \texttt{sawtooth}   & -1.042 & -241 (73\%) & -1.169 (79\%)          & -928 & -1.135 (74\%) & -1.201 (82\%)      & -1.024 & -1.144 (84\%) & -1.218 (88\%)    \\
        \texttt{square}     & -1.109 & -1.211 (89\%) & -1.225 (97\%)        & -947 & -1.090 (68\%) & -1.185 (91\%)      & -1.019 & -1.188 (82\%) & -1.216 (88\%)    \\
        \hline
    \end{tabular}}
\end{table*}} %

\begin{figure*}
    \centering
    \resizebox{\textwidth}{!}{\input{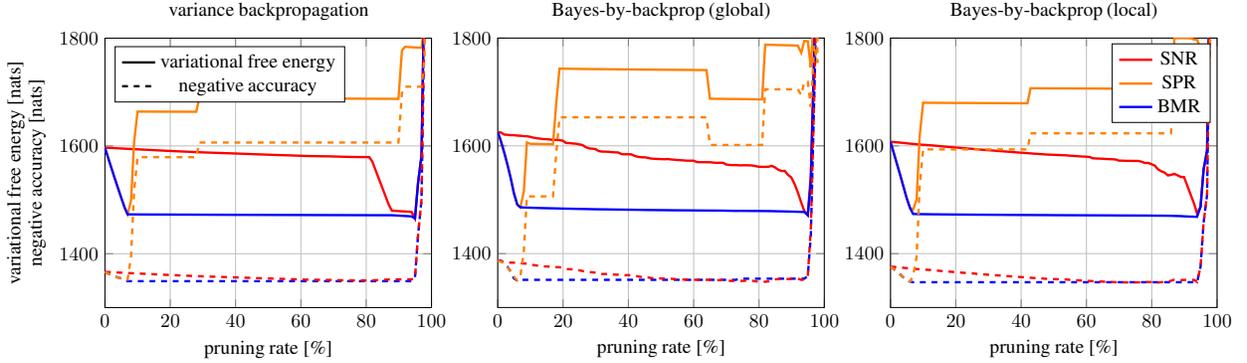}}
    \vspace*{-3mm}\caption{Comparison of the signal-to-noise ratio (SNR), signal-plus-robustness (SPR) and Bayesian model reduction (BMR) pruning metrics on the \texttt{boston} dataset. The different plots illustrate the different inference algorithms: variance backpropagation \cite{haussmann_sampling-free_2019} and Bayes-by-backprop \cite{blundell_weight_2015} with global \cite{kingma_auto-encoding_2014} and local \cite{kingma_variational_2015} reparameterization. Solid lines corresponds to the variational free energy and dashed lines to the negative accuracy from \eqref{eq:VFE-complexity}. Their difference corresponds to the complexity term in \eqref{eq:VFE-complexity}. The different colors denote the different pruning methods.}
    \label{fig:experiments:comparison}
\end{figure*}

\ifIEEE
    \subsection{COMPARISON TO STATE-OF-THE-ART} \label{sec:experiments:comparison}
\else
    \subsection{Comparison to state-of-the-art} \label{sec:experiments:comparison}
\fi
Comparison of our proposed algorithm against state-of-the-art methods, such as SNR and SPR, is not straightforward, because of their tunable pruning thresholds.
These methods require evaluating the performance of the model for different pruning rates in order to find the optimal pruned model.
Our method has a clear stopping criterion and therefore does not need to evaluate the model performance for different pruning rates, whereas SNR- and SPR-based pruning relies on a trial-and-error procedure for finding the optimal pruning rate.
Furthermore, extending SNR- and SPR-based pruning with retraining is ill-founded as their pruning and training objectives differ from each other.
Nonetheless, we perform a comparison of the different pruning methods (SNR, SPR and BMR) by evaluating the VFE for different pruning rates, without retraining.
Figure~\ref{fig:experiments:comparison} shows the capability of reducing the VFE for all pruning methods on the \texttt{boston} dataset.
Furthermore, the underlying negative accuracy as defined in \eqref{eq:VFE-complexity} is plotted to give insights in the individual VFE contributions.
Across all of our experiments, the minimum VFE value for SNR is observed to always be greater or equal to the minimum value for BMR.
The flat region in the BMR pruning metric corresponds to model parameters whose full posteriors are close to their full priors.
After training we observe an often significant group of parameters, whose sufficient statistics of the full posterior are close to those of the full prior.
These parameters are found not to affect the accuracy of the model significantly during training, and are therefore optimized to be close to the original prior to minimize model complexity.
As a result, the cluster of parameters located around the sufficient statistics of the full prior is pruned consecutively with corresponding $\Delta \mathrm{F}_\theta \approx 0$, leading to the plateaus in Figure~\ref{fig:experiments:comparison}.

From Figure~\ref{fig:experiments:comparison} it can be observed that both SNR and BMR reach  almost the same minimum at high pruning rates.
This can be attributed to the similarity in pruning curves for larger pruning thresholds as illustrated in Figure~\ref{fig:pruning-comparison}.
While SNR and BMR are both able to reach a certain minimum in a single pruning stage, we argue that BMR is the superior method in pruning BNNs.
A big advantage of BMR over SNR is the predefined pruning threshold.
SNR would require a search for the minimum VFE over different pruning thresholds, leading to a very expensive pruning method for models with a large number of model parameters.
BMR is able to reach the minimum in a single step as we only need to remove those parameters which decrease the VFE.
In addition, the objective of SNR has no relation to VFE and often in the literature different metrics are used to measure the effectiveness of SNR-based pruning~\cite{blundell_weight_2015}. 
Since the training and pruning of a BNN have a different objective, there is no guarantee that the pruned model will have a reached an optimum in terms of VFE.
With BMR the objectives are equal, indicating the better performance of BMR in minimizing the VFE through pruning.

\ifIEEE
    \section{RELATED WORK}\label{sec:related}
\else
    \section{Related Work}\label{sec:related}
\fi

In this section, we discuss related work corresponding to pruning in BNNs and compare these methods to our approach.
However, before doing so, we provide motivation for our commitment to variational inference in Section~\ref{sec:inference}.
    
\ifIEEE
    \subsection{PROBABILISTIC INFERENCE} \label{sec:related:inference}
\else
    \subsection{Probabilistic inference} \label{sec:related:inference}
\fi
The intractability of probabilistic inference in BNNs has produced two main groups of approximation methods: 1) Markov chain Monte Carlo (MCMC) sampling \cite{chib-1995, chapman-2011, hoffman2011nouturn}, which uses randomized sampling to approximate posterior distributions, and 2) variational inference, which minimizes a bound on Bayesian evidence to approximate the posterior distribution. In comparison to variational inference, MCMC methods scale poorly to large models with thousands or millions of parameters.
% \kk{Could you elaborate here a little bit what a large model is?}
Therefore, in this paper, we focused on variational inference methods for BNN training.

Variational inference for BNNs is well-represented in the literature \cite{mackay_practical_1992, blundell_weight_2015, hernandez-lobato_probabilistic_2015, soudry_expectation_2014, wu_deterministic_2019, haussmann_sampling-free_2019} with extensions to classification tasks \cite{ghosh_assumed_2016} and non-normal prior or variational posterior distributions \cite{louizos_bayesian_2017, sun_learning_2017, blundell_weight_2015}.
Throughout the experiments of Section~\ref{sec:experiments} we validate our pruning approach for two of the most common variational inference algorithms: Bayes-by-backprop \cite{blundell_weight_2015} and variance backpropagation \cite{haussmann_sampling-free_2019}.
Bayes-by-backprop \cite{blundell_weight_2015} approximates the accuracy term in \eqref{eq:VFE-complexity} by sampling the output with different possible parameterizations \cite{kingma_auto-encoding_2014, kingma_variational_2015}.
The variance backpropagation algorithm \cite{haussmann_sampling-free_2019} propagates moments through the network allowing for a closed-form approximation of the accuracy term in the VFE.

\ifIEEE
    \subsection{MODEL COMPRESSION} \label{sec:related:compression}
\else
    \subsection{Model compression} \label{sec:related:compression}
\fi
Recently, multiple survey papers have been published on the state-of-the-art compression methods for regular (non-Bayesian) neural networks \cite{mishra_survey_2020, cheng_survey_2020, zhang_compacting_2021}, which can be grouped into three main categories: network reduction, network modification and knowledge distillation.
This paper focuses specifically on network reduction.

The general aim of network reduction is to reduce the storage size or computational load of the network without changing its structure.
Within network reduction we can again distinguish between three subcategories: network pruning, quantization and low-rank decomposition.
The key idea of network pruning is to remove unimportant parts of the network such as parameters, nodes, layers and channels.
Components are removed based on certain heuristics and afterward the model is often retrained to regain accuracy.
The first and most common heuristic for pruning BNNs is the SNR \cite{graves_practical_2011, blundell_weight_2015}, defined as
\begin{equation} \label{eq:SNR}
    \mathrm{SNR}(\theta) =  \frac{\left\vert \mathbb{E}_{q(\theta)}[\theta]\right\vert }{\sqrt{\mathrm{Var}_{q(\theta)}[\theta]}},
\end{equation}
with $\theta$ being a parameter of the model.
The central moments are computed with respect to the corresponding (often approximate) posterior distribution $q(\theta)$.
SNR is then used as a ranking function where parameters with low values of SNR are pruned first.
The second commonly used heuristic is SPR from \cite[Ch. 4]{nalisnick_priors_2018}, defined as
\begin{equation} \label{eq:SPR}
    \mathrm{SPR}(\theta) = \left\vert\mathbb{E}_{q(\theta)}[\theta]\right\vert + \sqrt{\mathrm{Var}_{q(\theta)}[\theta]},
\end{equation}
which is used similarly as a ranking function, where parameters with low SPR values are pruned first.
\cite[Ch. 4]{nalisnick_priors_2018} argues that model parameters with large means and variances should not be pruned, instead of parameters with a large SNR.
A downside of both approaches is that they lack a clear stopping criterion: the user has to specify what values of SNR or SPR still are acceptable.
Choosing these thresholds is a time-consuming procedure that limits its automatability.

Next to these parameter pruning techniques, there exist alternative Bayesian compression methods.
In \cite{molchanov_variational_2017}, variational dropout was adopted to sparsify BNNs during training.
Earlier work of \cite{kingma_variational_2015} was extended to allow for learning dropout rates for individual weights.
\cite{louizos_bayesian_2017} uses hierarchical priors to prune nodes instead of weights in the network, which encourage the posterior to be sparse.
Additionally, the posterior uncertainty of the weights has been used to determine the optimal fixed-point precision to quantize them.
Similarly, the novel method of Bayesian automatic model compression (BAMC) \cite{wang_bayesian_2020} uses Dirichlet process mixtures to infer the optimal layer-wise quantization bit-width.
\ifIEEE
    \section{DISCUSSION AND FUTURE WORK}\label{sec:discussion}
\else
    \section{Discussion and future work}\label{sec:discussion}
\fi

The proposed pruning algorithm for BNNs has been shown to be effective in pruning a significant number of parameters.
Although this has led to the compression of the model, the currently unstructured pruning of model parameters might result in sparse structures that limit the gained computational efficiency.
For example, sparse matrices are often saved in a special data format that could limit the improved efficiency after pruning.
Instead, it would be more efficient to prune entire columns in these matrices at once or to prune entire kernels in convolutional layers.
Following up on this comment, investigation of structured prior and variational posterior distributions could lead to significant performance improvements.
Instead of assuming a full factorization between parameters, the covariance between parameters can be retained using multivariate distributions.
These structured pruning approaches are straightforward extensions of our approach, where the VFE difference is evaluated for the pruning of larger structured sets of parameters.
Yet we deem this extension an important direction for future research.

In the current experiments, we chose normal prior and variational posterior distributions, similarly to most of the works in the literature.
However, simultaneously there is significant effort being invested in selecting performance-improving prior distributions \cite{nalisnick_priors_2018, fortuin_priors_2021, fortuin_bayesian_2021}.
It would be interesting to investigate our pruning algorithm for different prior and variational posterior distributions, such as Laplace distributions, which enjoy a sparsifying behavior already.

In our work, we aimed at providing a proper, grounded, Bayesian framework for training and pruning BNNs. Therefore, we focused on the definition of our model and the appropriate inference methods. In line with this Bayesian view, we did not include any comparisons to the method of Variational Dropout~\cite{kingma_variational_2015,molchanov_variational_2017}. As presented in~\cite{hron_variational_2018}, there are two main issues with Variational Dropout; Being 1) the use of improper priors and 2) the singular approximating distributions of the posteriors. \cite{hron_variational_2018} states that Variational Dropout is not an approximation of variational inference and should therefore be treated as a non-Bayesian method. As a result, we do not provide a comparison to this approach and only compare to post-hoc pruning schemes. These can namely be applied on already trained networks, whereas variational dropout requires adaptation of existing models.

Finally, the current paper focuses on pruning the BNN model in a greedy and iterative fashion.
This process could yield suboptimal performance as parameters might be pruned prematurely due to the approximations in \eqref{eq:bmr-q}.
It would therefore be interesting to investigate the effects of reverting the pruning of qualifying variables in Algorithm~\ref{alg:bmr}.
This effective expansion of models would lead to a breakthrough in BNN design.
Models can then prune and grow themselves, depending on the complexity of the task.
A strategy could be to start out with prior distributions representing pruned parameters and then use BMR to evaluate the VFE improvement compared to the case where they had not been pruned.
The true power of this approach is obtained by adhering to the Bayesian approach and focusing on continual in-the-field learning.
Intertwining our pruning (and possible extension) procedure with this continual learning approach could lead to models that build themselves, automating costly trial-and-error model refinement tasks.
\ifIEEE
    \section{CONCLUSION}\label{sec:conclusion}
\else
    \section{Conclusion}\label{sec:conclusion}
\fi
This paper has extended the use of Bayesian model reduction to Bayesian neural networks.
We have introduced a principled parameter pruning approach for Bayesian neural networks, based on an iterative pruning-training scheme.
Importantly, this approach minimizes the same objective as during the training procedure  and has an intuitive threshold-free stopping criterion.
Because it carries the same optimization objective, retraining the network after pruning is well-grounded and can effectively alleviate discrepancies originating from the variational approximations.
Our results show that this algorithm achieves a lower variational free energy during pruning in comparison to state-of-the-art signal-to-noise ratio and signal-plus-robustness pruning.
Our pruning approach eases the development of neural networks for signal processing tasks in storage- and power-constrained devices.

\section*{Acknowledgments}
This work was partly financed by GN Hearing A/S. 
The authors would like to thank the \href{https://biaslab.github.io/}{BIASlab} team members and colleagues at GN Hearing for various insightful discussions related to this work.

\bibliographystyle{IEEEtran}
\bibliography{zotero, refs}

\appendix
\newpage
\ifIEEE
    \section{BAYESIAN MODEL REDUCTION DERIVATIONS}\label{app:bmr}
\else
    \section{Bayesian model reduction derivations}\label{app:bmr}
\fi
Bayesian model reduction \cite{friston_post_2011, friston_bayesian_2018} allows for the efficient recomputation of the posterior distribution and model evidence upon changes in the original prior distribution.
BMR assumes the following two probabilistic models
\begin{subequations}
    \begin{equation} \label{eq:app:bmr:model:full}
        p(Y, \bm{\theta}) = p(Y \midi \bm{\theta}) \, p(\bm{\theta}),
    \end{equation}
    \begin{equation} \label{eq:app:bmr:model:reduced}
        \tilde{p}(Y, \bm{\theta}) = p(Y \midi \bm{\theta}) \, \tilde{p}(\bm{\theta}).
    \end{equation}
\end{subequations}
Both models share the same likelihood model $p(Y \midi \bm{\theta})$ and differ through their prior distributions $p(\bm{\theta})$ and $\tilde{p}(\bm{\theta})$.
To distinguish between both situations, we use the $\tilde{\cdot}$ accent to represent the new or reduced \cite{friston_bayesian_2018} situation.

Suppose that we start out with the probabilistic model of \eqref{eq:app:bmr:model:full} and have computed the corresponding posterior distribution $p(\bm{\theta} \midi Y)$ and model evidence $p(Y)$.
It might be possible that the initial prior distribution $p(\bm{\theta})$ is not optimal and that we would like to reevaluate the computed terms for an alternative prior distribution $\tilde{p}(\bm{\theta})$.
However, performing all computations again can be a costly operation.
Therefore it would be ideal to leverage the information from the past to efficiently compute the posterior distribution $\tilde{p}(\bm{\theta}\midi Y)$ and model evidence $\tilde{p}(Y)$ of the new situation.

Based on the posterior distribution $p(\bm{\theta} \midi Y)$ and evidence $p(Y)$ we can compute the new posterior distribution $\tilde{p}(\bm{\theta}\midi Y)$ and new evidence $\tilde{p}(Y)$ for the new prior distribution $\tilde{p}(\bm{\theta})$ as follows.
First, we obtain the definitions for the original and new posterior distributions using Bayes' rule as
\begin{equation} \label{eq:app:bmr:bayes:full}
    p(\bm{\theta}\midi Y) = \frac{p(Y\midi \bm{\theta}) \, p(\bm{\theta})}{p(Y)},
\end{equation}
%     \begin{equation}\label{eq:app:bmr:bayes:reduced}
%         \tilde{p}(\bm{\theta}\midi Y) = \frac{p(Y\midi \bm{\theta}) \, \tilde{p}(\bm{\theta})}{\tilde{p}(Y)}.
%     \end{equation}
% \end{subequations}
and similarly for $\tilde{p}(\bm{\theta}\midi Y)$.
Both terms share the same likelihood $p(Y \midi \bm{\theta})$.
Rearranging both $p(\bm{\theta}\midi Y)$ and $\tilde{p}(\bm{\theta}\midi Y)$ gives
\begin{equation}
    \frac{p(\bm{\theta}\midi Y)p(Y)}{p(\bm{\theta})} = p(Y\midi\bm{\theta}) = \frac{\tilde{p}(\bm{\theta}\midi Y)\tilde{p}(Y)}{\tilde{p}(\bm{\theta})}.
\end{equation}
Here we can rearrange the expression to give the expression of the new posterior distribution
\begin{equation} \label{eq:app:bmr:result:posterior}
    \tilde{p}(\bm{\theta}\midi Y) = p(\bm{\theta}\midi Y) \frac{\tilde{p}(\bm{\theta})}{p(\bm{\theta})} \frac{p(Y)}{\tilde{p}(Y)},
\end{equation}
where only the term $\tilde{p}(Y)$ is unknown.
This term can be obtained by integrating both sides over $\bm{\theta}$ we obtain
\begin{equation}
    1 = \frac{p(Y)}{\tilde{p}(Y)} \int p(\bm{\theta} \midi Y) \frac{\tilde{p}(\bm{\theta})}{p(\bm{\theta})} \, \mathrm{d}\bm{\theta},
\end{equation}
from which we can obtain \eqref{eq:bmr:p-evidence} by rearranging and by taking the logarithm of both sides of the equality.
By substituting this result into \eqref{eq:app:bmr:result:posterior} leads to the new posterior distribution.
\ifIEEE
    \section{CLOSED-FORM EXPRESSION OF \eqref{eq:bmr-q} FOR NORMALS}\label{app:bmr-gaussian}
\else
    \section{Closed-form expression of \texorpdfstring{Eq.~\eqref{eq:bmr-q-vfe}}{EQ. (10b)} for normals}\label{app:bmr-gaussian}
\fi
Given the prior distribution $p(\theta) = \mathcal{N}(\theta \midi \mu_p, \sigma^2_p)$, posterior $q(\theta) = \mathcal{N}(\theta\midi \mu_q, \sigma^2_q)$ and reduced prior $\tilde{p}(\theta) = \mathcal{N}(\theta \midi \tilde{\mu}_p, \tilde{\sigma}^2_p)$. The reduced posterior distribution in \eqref{eq:bmr-q-posterior} can be derived as $\tilde{q}(\theta) = \mathcal{N}(\theta\midi \tilde{\mu}_q, \tilde{\sigma}^2_q)$, where
\begin{subequations}
    \begin{equation}
        \tilde{\sigma}^2_q = \left(\frac{1}{\sigma_q^2} + \frac{1}{\tilde{\sigma}_p^2} - \frac{1}{\sigma_p^2}\right)^{-1},
    \end{equation}
    \begin{equation}
        \tilde{\mu}_q = \tilde{\sigma}^2_q\left(\frac{\mu_q}{\sigma_q^2} + \frac{\tilde{\mu}_p}{\tilde{\sigma}_p^2} - \frac{\mu_p}{\sigma_p^2}\right).
    \end{equation}
\end{subequations}
The change in VFE $\Delta \mathrm{F}_\theta$ can be derived as
\begin{equation}
    \Delta \mathrm{F}_\theta = \frac{1}{2}\ln\left(\frac{\tilde{\sigma}_q^2 \sigma^2_p}{\tilde{\sigma}_p^2 \sigma^2_q}\right) - \frac{1}{2}\left(\frac{\mu_q^2}{\sigma_q^2} + \frac{\tilde{\mu}^2_p}{\tilde{\sigma}_p^2} - \frac{\mu_p^2}{\sigma_p^2} - \frac{\tilde{\mu}_q^2}{\tilde{\sigma}^2_q}\right).
\end{equation}

% All except for the reduced posterior distribution $\tilde{q}$ are either defined or obtained during training. 
% In the following equations, we will be using precisions next to variances, where $\lambda^{-1} = \sigma^2$. 
% We now can define the reduced posterior precision and mean as
% \begin{subequations}
%     \begin{equation} \label{eq:app:bmr-gaussian:red-post:variance}
%         \tilde{\lambda}_q = \lambda_q + \tilde{\lambda}_p - \lambda_p ,
%     \end{equation}
%     \begin{equation} \label{eq:app:bmr-gaussian:red-post:mean}
%         \tilde{\mu}_q = \tilde{\sigma}^2_q \left( \lambda_q \mu_q + \tilde{\lambda}_p \tilde{\mu}_p - \lambda_p \mu_p \right) .
%     \end{equation}
% \end{subequations}
% Given these parameters, we have the following analytical solution to the change in VFE according to Eq.~\eqref{eq:bmr-q-vfe}
% \ifIEEE
%     \begin{multline}
%         \Delta F = \frac{1}{2} \ln \frac{|\tilde{\lambda}_p| |\lambda_q|}{|\tilde{\lambda}_q| |\lambda_p|} - \\ \frac{1}{2} \left( \lambda_q \mu^2_q + \tilde{\lambda}_p \tilde{\mu}^2_p - \lambda_p \mu^2_p - \tilde{\lambda}_q \tilde{\mu}^2_q \right) .
%     \end{multline}
% \else
%     \begin{equation}
%         \Delta F = \frac{1}{2} \ln \frac{|\tilde{\lambda}_p| |\lambda_q|}{|\tilde{\lambda}_q| |\lambda_p|} - \\ \frac{1}{2} \left( \lambda_q \mu^2_q + \tilde{\lambda}_p \tilde{\mu}^2_p - \lambda_p \mu^2_p - \tilde{\lambda}_q \tilde{\mu}^2_q \right) .
%     \end{equation}
% \fi
\end{document}